\documentclass{article}

\usepackage{arxiv}

\usepackage[utf8]{inputenc} 
\usepackage[T1]{fontenc}    
\usepackage{hyperref}       
\usepackage{url}            
\usepackage{booktabs}       
\usepackage{amsfonts}       
\usepackage{nicefrac}       
\usepackage{microtype}      
\usepackage{lipsum}
\usepackage{graphicx}
\usepackage{amsmath}
\graphicspath{ {./images/} }

\title{HuDEx: Integrating Hallucination Detection and Explainability for Enhancing the Reliability of LLM responses}

\author{
 Sujeong Lee \\
Inha University\\
  Incheon, 22212, Republic of Korea \\
  \texttt{tnwjd025611@inha.edu} \\
   \And
 Hayoung Lee \\
  Inha University\\
  Incheon, 22212, Republic of Korea \\
    \texttt{gkdud000123@gmail.com} \\
   \And
  Seongsoo Heo \\
  Inha University\\
  Incheon, 22212, Republic of Korea \\
      \texttt{woo555813@inha.edu} \\
   \And
  Wonik Choi \\
  Inha University\\
  Incheon, 22212, Republic of Korea \\
  \texttt{wichoi@inha.ac.kr} \\}

\begin{document}
\maketitle
\begin{abstract}
Recent advances in large language models (LLMs) have shown promising improvements, often surpassing existing methods across a wide range of downstream tasks in natural language processing. However, these models still face challenges, which may hinder their practical applicability. For example, the phenomenon of hallucination is known to compromise the reliability of LLMs, especially in fields that demand high factual precision. Current benchmarks primarily focus on hallucination detection and factuality evaluation but do not extend beyond identification. This paper proposes an explanation enhanced hallucination-detection model, coined as HuDEx, aimed at enhancing the reliability of LLM-generated responses by both detecting hallucinations and providing detailed explanations. The proposed model provides a novel approach to integrate detection with explanations, and enable both users and the LLM itself to understand and reduce errors. Our measurement results demonstrate that the proposed model surpasses larger LLMs, such as Llama3 70B and GPT-4, in hallucination detection accuracy, while maintaining reliable explanations. Furthermore, the proposed model performs well in both zero-shot and other test environments, showcasing its adaptability across diverse benchmark datasets. The proposed approach further enhances the hallucination detection research by introducing a novel approach to integrating interpretability with hallucination detection, which further enhances the performance and reliability of evaluating hallucinations in language models.
\end{abstract}

\section{Introduction}
Recent advancements in large language models (LLMs) have showcased their potential in natural language processing (NLP) \cite{Bubeck2023}. While LLMs can generate effective responses across diverse tasks, they are also limited by certain critical issues. One such limitation is hallucination, where the model produces information that is factually incorrect or generates content not requested or instructed by the user. This problem can lead to the spread of incorrect information, particularly problematic in fields where accuracy and reliability are crucial, thereby limiting the applicability of LLMs in various industries. Consequently, hallucination is a major issue undermining the reliability of LLMs, prompting significant research into solutions.

Recent studies have focused on developing benchmarks to detect and evaluate hallucinations and methods for mitigating them. For example, FELM \cite{Chen2023} provides a benchmark for assessing the factuality of LLMs by identifying factual errors in response segments through text-segment-based annotations. TruthfulQA \cite{Lin2022} evaluates whether language models produce truthful responses, aiming to detect non-factual responses across various domains. Similarly, QAFactEval \cite{Fabbri2022} proposes a QA-based metric for assessing factual consistency in summarization tasks, effectively detecting and evaluating factual errors.

However, these studies primarily focus on evaluating or detecting hallucinations or a lack of factual inaccuracies, rather than actively improving the model's reliability. This limitation underscores the need for approaches that not only assess factual errors but also actively contribute to improving the quality of model responses. Additionally, benchmark-based evaluation methods may struggle with the real-time detection of hallucinations in model-generated responses.

To address these gaps, we propose a specialized model named HuDex designed to detect hallucinations in LLM responses and provide detailed explanations of these hallucinations. Unlike existing benchmarks, our model not only identifies hallucinations but also offers specific explanations, helping users understand the model’s output and assisting the model in refining its responses. This approach aims to improve the reliability of LLM responses. 

\begin{figure}[h]
\centering
\includegraphics[width=0.5\textwidth]{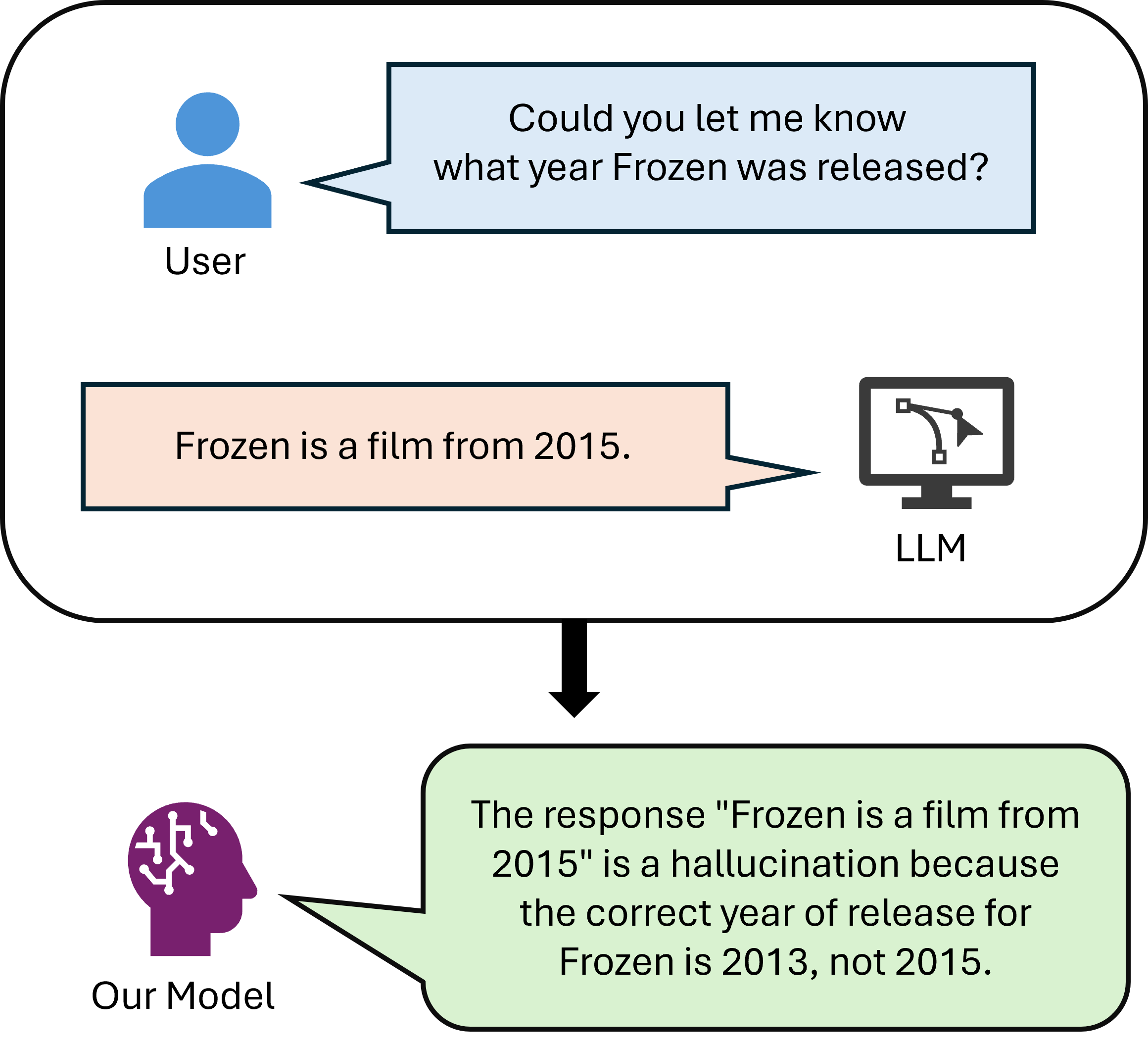}
\caption{Schematic Representation of Our Hallucination Model
}\label{fig1}
\end{figure}

The key contributions of our proposed model are: 

1. Moving beyond standardized hallucination benchmarks, the proposed model enables proactive detection despite its smaller size.

2. By providing detailed explanations of detected hallucinations, the model enhances user understanding and contributes to the improvement of model performance.

3. Through an analysis focused on hallucinations, a more nuanced evaluation of the hallucination domain is possible compared to general-purpose LLMs, and this can be effectively used to evaluate other LLMs.

\section{Related Work}
\label{sec:relatedwork}

\subsection{Definitions of Large Language Models}
A Large Language Model is an artificial intelligence model based on the Transformer architecture \cite{Vaswani2017}. It refers to a pre-trained language model (PLM) with a parameter size exceeding a certain threshold \cite{Zhao2023}. LLMs are trained on massive datasets and typically have billions to hundreds of billions of parameters. Due to the extensive data used in their training, LLMs exhibit exceptional performance across various NLP tasks, including text generation, translation, and summarization.

Notably, LLMs that surpass a certain parameter scale demonstrate emergent abilities not found in smaller models. Examples of these abilities include in-context learning, instruction following, and chain-of-thought (CoT) reasoning \cite{Wei2022}. These capabilities enable LLMs to handle more complex tasks, such as advanced reasoning, problem-solving, and generating multi-turn responses.

Although LLMs are primarily used for general downstream tasks, their increasing significance in both academia and industry has led to research into domain-specific LLMs. Examples include the Med-PaLM series for the medical domain \cite{Singhal2023} and FinGPT for the financial domain \cite{Yang2023}. These advancements underscore the growing need for LLMs not only in language generation but also in addressing specialized tasks across various fields.

\subsection{Definitions of Hallucination}
In NLP, hallucination refers to content that is unreliable or illogical compared to the provided source material \cite{Ji2023}, \cite{Maynez2020}. Previous studies categorize hallucinations into two broad two types: intrinsic and extrinsic \cite{Ji2023}, \cite{Maynez2020},\cite{Dziri2021}, \cite{Huang2023a}.

Intrinsic hallucination occurs when the generated output contradicts the source content. For example, this happens when a model produces information that conflicts with the given data in response to a factual question. In contrast, extrinsic hallucinations involve outputs that include unverifiable or nonexistent information. This often occurs when the model generates content that cannot be corroborated by the source material.

In the context of LLMs, hallucination can be defined more specifically. LLM hallucinations, which prioritize user instructions and interactions, can be categorized based on factuality and faithfulness \cite{Huang2023b}. Factual hallucinations arise when a model generates outputs that are based on real-world information but are either incorrect or unverifiable. For instance, if the model inaccurately presents well-known facts or mentions nonexistent information, it is considered a factual hallucination. Faithfulness-related hallucinations occur when the model generates responses unrelated to user instructions or the provided content, or when it produces internally inconsistent answers. This type of hallucination is particularly important in conversational models.

The issue of hallucination may stem from several factors, including the use of outdated data during the data collection process \cite{Kasai2024} or biased data \cite{Bender2021} used for model training \cite{Huang2023b},\cite{chiang2022overcomingtheoreticallimitationselfattention},  \cite{Li2023a}. Furthermore, the risk of hallucinations tends to increase with the size and complexity of the models.

\subsection{LLM-Based Evaluation of LLMs}
One of the key challenges discussed alongside the development of LLMs is the difficulty in accurately evaluating the context and meaning of generated responses using traditional quantitative metrics. While human evaluation has been employed to address this limitation, it has considerable drawbacks, particularly in terms of time and resource consumption \cite{Bubeck2023},\cite{Bang2023}.

To overcome these challenges, the use of LLMs as evaluation tools, or “LLM judges,” has gained attention. \cite{Zheng2024} pioneered an LLM-based evaluation framework, showing that strong LLMs achieved over 80\% agreement with human experts in evaluations. Subsequent studies by \cite{Bai2023}, \cite{Liu2023}, and \cite{Li2023b} have expanded on this approach, further validating the utility of LLM judges.

The introduction of LLM judges provides an efficient solution for evaluating large-scale data, where human evaluation may be impractical. In addition to quantitative  assessments, LLM judges offer qualitative evaluations based on their understanding of context and adherence to user instructions, making them versatile tools for comprehensive evaluation.   

\section{Data Construction}
\label{sec:dataconstruction}

\subsection{Datasets}
For training, we utilized the HaluEval, FactCHD, and FaithDial datasets, as summarized in Table 1.

The HaluEval dataset \cite{Li2023c} is a hallucination evaluation benchmark designed to assess the likelihood of hallucinations based on content type. It consists of 30,000 examples across three tasks: question answering, knowledge-based dialogues, and text summarization, along with 5,000 general user queries that include ChatGPT responses. In this study, we used the question-answering and knowledge-based dialogue subsets as training data. Both subsets focus on detecting hallucinations based on provided knowledge, allowing the model to learn how to identify intrinsic hallucinations.

The FactCHD dataset \cite{Chen2024} is a benchmark specifically designed to detect hallucinations that conflict with factual information in LLMs. It evaluates factual accuracy in the context of a wide range of queries and responses, facilitating factual reasoning during evaluation. Unlike HaluEval, the FactCHD dataset does not include a pre-existing knowledge base, enabling the model to learn to detect hallucinations in scenarios with limited reference material.

The FaithDial dataset \cite{Dziri2022a} is designed to minimize hallucinations and improve the accuracy of information-seeking dialogues. It was built by modifying the Wizard of Wikipedia (WOW) benchmark to include hallucinated responses. The dataset includes a BEGIN \cite{Dziri2022b} label that categorizes responses based on their relationship to the knowledge source and their contribution to the dialogue. For binary classification of hallucination detection, we preprocessed the dataset by excluding the Generic and Uncooperative categories. Additionally, since each data point includes both a response and an original response, we split them into two distinct responses. This allowed us to create two separate data instances with the same knowledge and dialogue context but different responses, thereby augmenting the training data. 

\begin{table}[h]
\centering
\caption{Dataset Information}
\begin{tabular}{l c c}
\hline
$\text{Dataset}$& $\text{Train}$& $\text{Test}$\\
\hline
$\text{HaluEval Dialogue}$& 9,000 & 1,000 \\
\hline
$\text{HaluEval QA}$& 9,000 & 1,000 \\
\hline
$\text{FaithDial}$& 18,357 & 3,539 \\
\hline
$\text{FactCHD}$& 51,838 & 6,960 \\
\hline
\end{tabular}
\end{table}

\subsection{Explanation Generation}
The primary goal of our model is not only to detect hallucinations in generated responses but also to provide explanations for the reasoning behind these judgments. A simple example of this process is illustrated in Figure 1. To achieve this, the model must be trained on explanation data. While the FactCHD dataset includes explanations, the HaluEval and FaithDial datasets do not. Therefore, we used the Llama3 70B \cite{Team2024} model to generate explanation data for hallucination detection in the HaluEval and FaithDial datasets.

During the explanation generation process, we also generated answers corresponding to the hallucination labels. This step ensured that the hallucination labels predicted by the model during explanation generation aligned with the existing hallucination labels in HaluEval and FaithDial datasets.

Upon analyzing the model’s predictions, we found that 0.5\% of the responses failed to understand the prompt and asked for clarification, and 4.2\% were classified as anomalies. Excluding these cases, 95.3\% of the responses adhered to the expected format. As shown in Table 2, the accuracy of valid responses was 98.3\%. Ultimately, 93.7\% of the hallucination labels from HaluEval and FaithDial matched the model’s predicted answers, and only the verified matching data were used for training. 

To further assess the quality of the generated explanations, we conducted statistical sampling. We defined the population as the set of generated explanations, with a confidence level of 99\%, a conservatively estimated defect rate of p = 0.5, and a margin of error set at 2\%. Through human evaluation of the selected sample, we validated the explanations to ensure the accuracy and relevance of the reasoning provided.

\begin{table}[h]
\centering
\caption{Confusion Matrix of Model Predictions VS Actual Answers (Proportional Representation)}
\begin{tabular}{|c|c|c|}
\hline
& $\begin{array}{c}
\text{Actual} 
\text{Positive}
\end{array}$& $\begin{array}{c}
\text{Actual} 
\text{Negative}
\end{array}$\\
\hline
$\text{Predicted Positive}$& $52.0\%$& $1.7\%$\\
\hline
$\text{Predicted Negative}$& $0\%$& $46.3\%$\\
\hline
\end{tabular}
\end{table}

\section{Model Training and Inference}
\label{sec:modeltrainingandinference}

\subsection{Training}
We used the Llama 3.1 8B model \cite{Team2024} for training and applied low-rank adaptation (LoRA) \cite{Hu2022}, a method under parameter efficient tuning (PEFT). The task prompts for training were divided into two main categories: hallucination detection and hallucination explanation. The model was trained on both tasks using the same dataset.

\subsection{Inference}
The prompt structure for inference focuses on two key elements: persona provision and task stage provision. Persona provision ensures that the model understands the specific task’s goal before generating responses, encouraging deeper analysis of the given information. By defining the task’s context and role in advance, we aim for more consistent outputs. To generate a persona, we provided ChatGPT with task details and received recommendations for suitable persona candidates. After a human filtering process, we selected a hallucination expert persona to detect hallucinations.

\begin{figure}[h]
\centering
\includegraphics[width=0.9\textwidth]{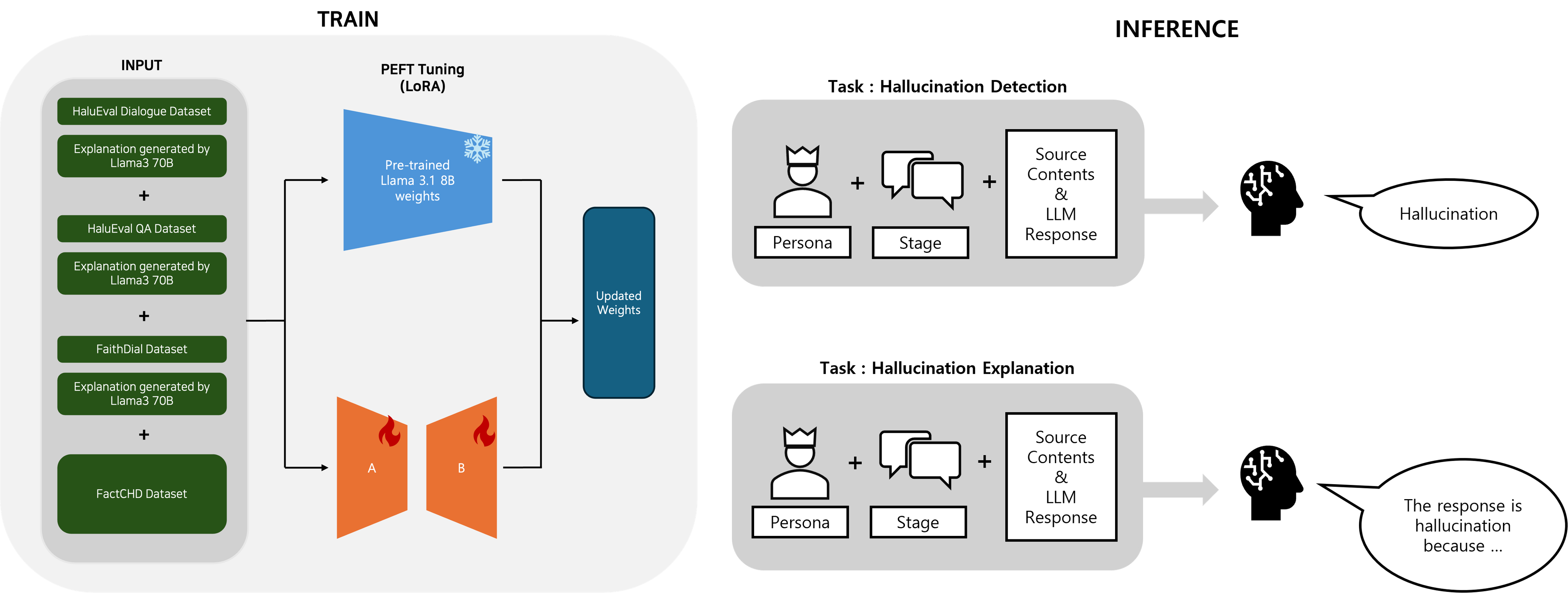}
\caption{Overview of HuDEx: Training and Inference}\label{fig2}
\end{figure}

Task stage provision guides the model to approach complex problems systematically when generating responses. The prompt stages are structured adaptively based on the task and data characteristics. If background knowledge is available, the model generates responses based on it; otherwise, it relies on context and its inherent knowledge. The stage structure also varies depending on whether the task focuses on hallucination detection or explanation generation. Stages can be divided or combined based on the specific needs of each task.

An overview of the training and inference process can be found in Figure 2, and brief examples of both the stage and persona structures are shown in Figure 3.

\section{Experiments}
\label{sec:experiments}

\subsection{Datasets}
For the detection and explanation generation experiments, we used the test sets from HaluEval dialogue, HaluEval QA, FaithDial and FactCHD, which were also used during training. The HaluEval datasets, both for dialogue and QA tasks, provide background knowledge, so we applied inference prompts designed to incorporate this information. FaithDial also utilized the same inference prompt. For the FactCHD dataset, which does not include background knowledge, we used the inference prompt stages suited for tasks without background knowledge. The persona was consistently provided across all tasks, regardless of the presence or absence of background knowledge.

For zero-shot detection, we conducted experiments on HaluEval subsets not used during training, specifically HaluEval summarization and HaluEval general. The HaluEval summarization dataset focuses on detecting hallucinations in document summaries, while the HaluEval general dataset evaluates hallucination detection in ChatGPT responses to user queries. Since both datasets lack background knowledge, we used inference prompts designed for tasks without background knowledge.

\begin{figure}[h]
\centering
\includegraphics[width=0.9\textwidth]{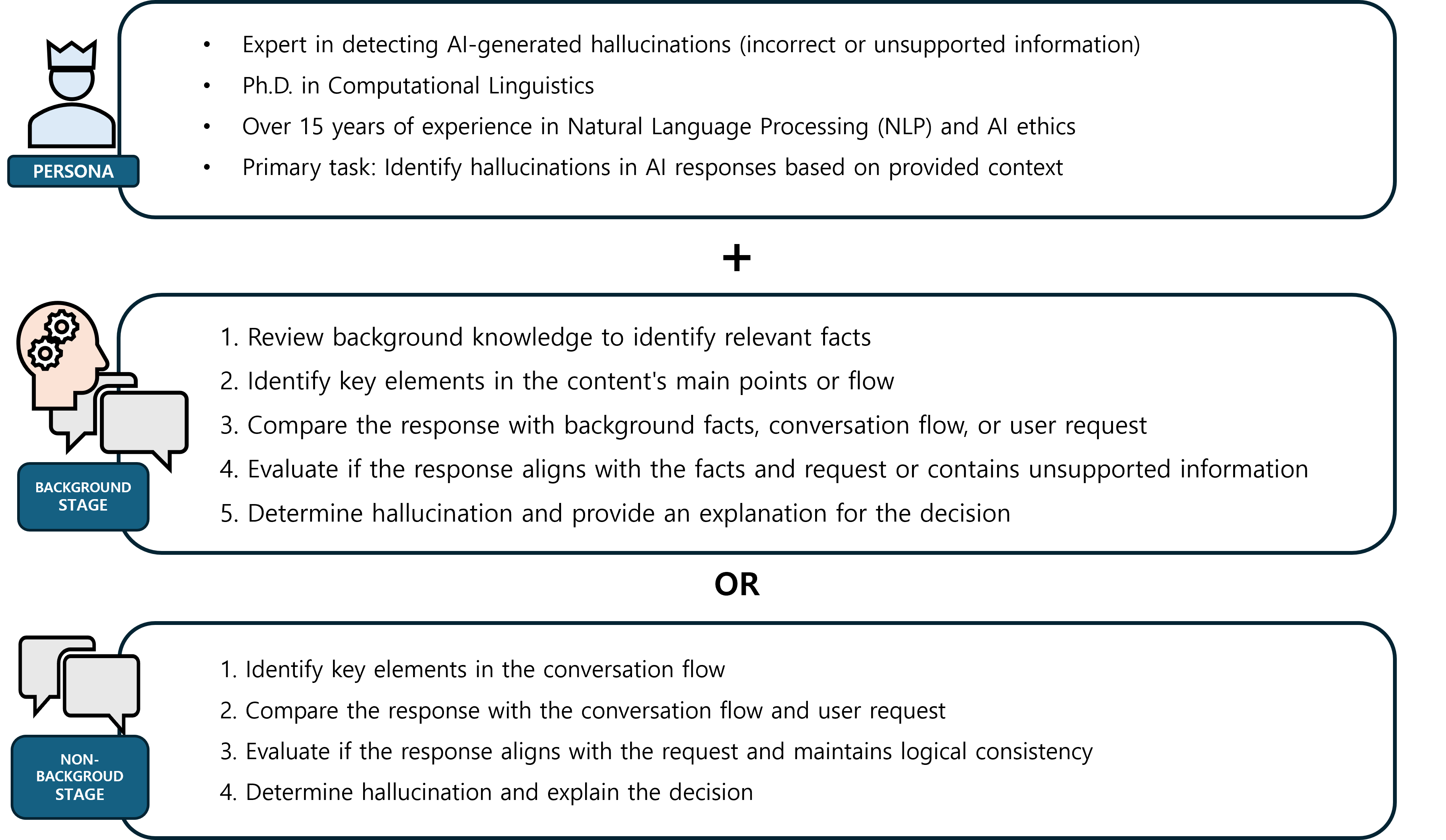}
\caption{Examples of Persona and Steps Used in Inference Prompts}\label{fig3}
\end{figure}

\subsection{Test Setting}
\subsubsection{Detection Experiments}
For the detection experiments, we compared our HuDEx to two LLMs, GPT-4 \cite{OpenAI2024} and Llama 3 70B. These models received the same inference prompts as our model and were tasked with classifying whether the responses contained hallucinations.

\subsubsection{Explanation Generation Experiments}
To evaluate the explanations generated by each model, we used an LLM judge and conducted main experiment. The experiment followed a single-answer grading approach, where each model’s response was individually scored. 

In the single-answer grading experiment, we divided the evaluation into two categories: factuality and clarity. Factuality assessed whether the explanation contained hallucinations, contradictions, or accurately reflected the given information. Clarity evaluated how clearly and thoroughly the reason was articulated. Each criterion was scored on a 3-point scale, with a maximum total score of 6 points.

We used GPT4o as the judge for experiment. In the HaluEval and FaithDial dataset, we compared the explanations generated by our model against those from Llama 3 70B, with GPT4o providing the final judgments. For the FactCHD dataset, we compared the explanations generated by HuDEx against the explanations included in the FactCHD dataset itself.

\section{Results}
\label{sec:results}

\subsection{Detection Results}
\subsubsection{Test Data Detection}

In this experiment, binary classification was performed to distinguish hallucinations from non-hallucinations using the test sets from the training data, with accuracy as the evaluation metric. Table 3 compares the performance of  Llama3 70B, GPT4o, and our model across benchmark datasets such as HaluEval dialogue, HaluEval QA, FactCHD, and FaithDial.

The experimental results show that our HuDEx outperformed the larger models, Llama3 70B and GPT4o, across all benchmarks. Specifically, it achieved an accuracy of 80.6\% on the HaluEval dialogue dataset, surpassing Llama3 70B (71.8\%) and GPT4o (72.5\%), indicating superior performance in detecting hallucinations in conversational response.

In the HaluEval QA dataset, our model again achieved the highest accuracy of 89.6\%, outperforming GPT4o (86.6\%) and Llama3 70B (82.7\%). This demonstrates its refined ability to detect hallucinations in QA tasks.

On the FactCHD and FaithDial datasets, HuDEx recorded accuracies of 70.3\% and 58.8\%, respectively, continuing to show strong performance on both benchmarks. On the FactCHD dataset, HuDEx outperformed Llama3 70B by \~11\%, confirming its effectiveness in hallucination detection even when background knowledge is unavailable. On the FaithDial dataset, our HuDEx also significantly outperformed GPT4o (50.6\%), achieving 58.8\%, which highlights its consistent performance on a different type of conversation-based dataset compared to HaluEval dialogue.

 These results demonstrate that our model consistently delivers superior performance in hallucination detection across various benchmark datasets, outperforming larger models.

 \begin{table}[h]
\centering
\caption{Test Data Detection Results (Accuracy)}
\begin{tabular}{lcccc}
\hline
$\text{Model}$& $\text{HaluEval Dialogue}$& $\text{HaluEval QA}$& $\text{FactCHD}$& $\text{FaithDial}$\\
\hline
$\text{Llama3 70B}$& 71.8 & 82.7 & 59.4 & 47.9 \\
\hline
$\text{GPT4o}$& 72.5 & 86.6 & 61.2 & 50.6 \\
\hline
$\text{HuDEx}$& $\mathbf{80.6}$& $\mathbf{89.6}$& $\mathbf{70.3}$& $\mathbf{58.8}$\\
\hline
\end{tabular}
\end{table}

\subsubsection{Zero-Shot Detection}
Table 4 presents the results of the binary classification experiment on hallucination vs. non-hallucination in a zero-shot setting. This experiment evaluated the model’s hallucination detection performance on unseen data using the HaluEval summarization and HaluEval general datasets, which were not included in the training data. Accuracy was used as the evaluation metric, consistent with the methodology in previous experiments.

On the HaluEval summarization dataset, HuDEx achieved an accuracy of 77.9\%, outperforming Llama3 70B (69.55\%) and GPT4o (61.9\%). This demonstrates the model’s ability to effectively detect hallucinations in summary texts of original content.

The HaluEval general dataset consists of queries posed by real users to GPT models, often containing complex responses that go beyond typical conversational text. This complexity makes hallucination detection more challenging and serves as an important benchmark for evaluating model reliability on unstructured data. On this dataset, GPT4o recorded the highest accuracy at 78.0\%, while our model achieved 72.6\%. These results suggest that while HuDEx delivers consistent performance on complex responses, there is still room for improvement.

\begin{table}[h]
\centering
\caption{Zero-shot data detection results (Accuracy)}
\begin{tabular}{lcc}
\hline

$\text{Model}$& $\begin{array}{c}
\text{HaluEval} 
\text{Summarization}
\end{array}$& $\begin{array}{c}
\text{HaluEval} 
\text{General}
\end{array}$\\
\hline
$\text{Llama3 70B}$& 69.55 & 76.2 \\
$\text{GPT4o}$& 61.9 & $\mathbf{78.0}$\\
$\text{HuDEx}$& $\mathbf{77.9}$& 72.6 \\
\hline
\end{tabular}
\end{table}

\subsection{Explanation Generation Results}
\subsubsection{Single-Answer Grading}
This experiment presents the evaluation of hallucination explanations generated by Llama3 70B and our model, as assessed by the LLM judge. The results, shown in Table 5, were obtained from the HaluEval dialogue, HaluEval QA, and FaithDial datasets. Explanations were evaluated based on two criteria: factuality and clarity, each scored out of 3 points, for a maximum combined score of 6 points.

When comparing the performance of Llama3 70B and our HuDEx in terms of factuality, Llama3 70B scored lower on the HaluEval dialogue dataset with 1.932 points but achieved relatively higher scores on HaluEval QA and FaithDial, with 2.416 and 2.587 points, respectively. In contrast, our model outperformed Llama3 70B on factuality for the HaluEval dialogue dataset, though it scored slightly lower on HaluEval QA (2.299) and FaithDial (2.216). Despite the variations in scores across datasets, HuDEx demonstrated consistent factual accuracy, indicating its ability to provide reliable information.

In terms of clarity, Llama3 70B achieved the highest score on the FaithDial dataset with 2.451 points, while our model closely followed with 2.417 points. On the HaluEval dialogue and HaluEval QA datasets, our model outperformed Llama3 70B, scoring 2.413 and 2.523 points, respectively. This indicates that HuDEx provides clearer and more easily understandable explanations for hallucinations.

Overall, our HuDEx demonstrated competitive performance in terms of factuality, clarity, and overall scores compared to Llama3 70B. These results support that our model consistently delivers reliable and clear hallucination explanations.

\begin{table}[h]
\centering
\caption{Comparison of Hallucination Explanations Between Llama3 70B and Proposed model (LLM Judge Evaluation)}
\begin{tabular}{llccc}
\hline
Model& Dataset& Factuality (3)& Clarity (3)& Overall (6)\\
\hline
Llama3 70B& HaluEval Dialogue & 1.932 & 2.302 & 4.256 \\
                   & HaluEval QA       & \textbf{2.416} & 2.153 & 4.569 \\
                   & FaithDial         & \textbf{2.587} & \textbf{2.451} & \textbf{5.038} \\
\hline
HuDEx& HaluEval Dialogue & \textbf{2.116} & \textbf{2.413} & \textbf{4.528} \\
                   & HaluEval QA       & 2.299 & \textbf{2.523} & \textbf{4.822} \\
                   & FaithDial         & 2.216 & 2.417 & 4.633 \\
\hline
\end{tabular}
\end{table}

The next experiment evaluated the original explanations from the FactCHD dataset against those generated by our model, with results shown in Table 6. The conversion ratio was used to compare the performance of our HuDEx as a percentage, with the FactCHD score serving as the maximum (100\%).

For factuality, FactCHD recorded a score of 2.2549, while our model scored slightly lower at 2.236. The conversion ratio for factuality was 99\%, indicating that although FactCHD’s original explanations had slightly higher factual accuracy, HuDEx performed very closely to this benchmark.

In terms of clarity, FactCHD achieved a score of 2.439, while our model scored slightly lower at 2.37. The conversion ratio for clarity was 97\%, suggesting that while our model’s explanations were marginally less clear than FactCHD’s, they remained highly comparable in clarity. In conclusion, HuDEx showed performance similar to FactCHD, with conversion ratios ranging from 97\% to 99\%. These results demonstrate that HuDEx generates explanations nearly equivalent in quality to the original explanations provided in the FactCHD dataset.

\begin{table}[h]
\centering
\caption{LLM Judge Evaluation of Explanations:FactCHD original vs HuDEx}
\begin{tabular}{lccc}
\hline
& $\text{Factuality (3)}$& $\text{Clarity (3)}$& $\text{Overall (6)}$\\
\hline
$\text{FactCHD}$& 2.2549 & 2.439 & 4.697 \\
\hline
$\text{HuDEx}$& 2.236 & 2.37 & 4.61 \\
\hline
\hline
$\text{Conversion Ratio}$& $99\%$& $97\%$& $98\%$\\
\hline
\end{tabular}
\end{table}

\section{Conclusion}
\label{sec:conclusion}

The hallucination phenomenon in large language models (LLMs) presents a significant challenge that needs to be addressed in practical applications. This study proposes a model called HuDEx specifically designed to detect hallucinations in LLM-generated responses and provide explanations for them. By offering such feedback, the model contributes to both user understanding and the improvement of LLM, fostering the generation and evaluation of more reliable responses.

However, a key limitation of the model is its reliance on the LLM’s inherent knowledge when sufficient source content is unavailable for detecting and explaining hallucinations. This dependency can reduce the clarity of the explanations and, in some cases, introduce hallucinations into the explanations themselves.

Despite this limitation, the study demonstrates strong potential for detecting and explaining hallucinations. Future research should focus on overcoming these challenges and exploring methods to improve the model’s performance. For example, integrating external knowledge retrieval systems could reduce the model’s reliance on its internal knowledge, while enhancing reasoning-based validation could lead to more reliable explanations.

Additionally, we aim to develop an automated feedback loop in future work. This system would allow for continuous correction and improvement of hallucinations, contributing to greater reliability and consistency in LLMs over time.

\bibliographystyle{unsrt}  

\bibliography{references}

\end{document}